\documentclass{article}

\usepackage[final]{corl_2026}

\usepackage{graphicx}
\usepackage{wrapfig}
\usepackage{booktabs}
\usepackage{multirow}
\usepackage{makecell}
\usepackage{float}
\usepackage[table]{xcolor}
\usepackage{comment}
\usepackage{amssymb, amsmath}
\usepackage{ulem}
\usepackage{enumitem}
\usepackage{silence}
\title{Does Your World Model Know \\ Actions Have Consequences?}

\author{
  Ayush Gaggar 
  \begingroup\thanks{correspondence to \texttt{agaggar@u.northwestern.edu}, \texttt{jaredberry2026@u.northwestern.edu}, and \texttt{t-murphey@northwestern.edu}.}\endgroup
  \quad Jared Berry \quad Todd D. Murphey\\
  Department of Mechanical Engineering\\
  Northwestern University, Evanston, IL, USA\\
}

\begin{document}
\maketitle


\vspace{-0.6cm}
\begin{abstract}
    Embodied world models (WMs) are implicit learning systems that predict the next future latent state directly from observations and actions. 
    Unique from other end-to-end learning systems, the dynamics predictor of WMs can simulate the environment's response to alternative actions without physical deployment or pre-specifying states of the dynamical system.
    Current metrics focus on the quality of model outputs, e.g., clarity of reconstructions or task success, but fail to assess the fidelity of modeling consequences of counterfactual actions.
    Inspecting the forced response to action perturbations through sensitivity analysis is one way to \textit{audit} how well WMs can predict the consequences of counterfactual actions.
    Deploying WMs on hardware will require new evaluation frameworks to analyze the tradeoffs between state error and accurate sensitivity prediction.
\end{abstract}

\section{Introduction}
\label{intro}
\vspace{-0.3cm}

``World models" (WMs) have become an increasingly overloaded term, and refer to any combination of forecasting models, dynamics predictor models, latent-imagination based planners, video predictors, and synthetic data generators ~\citep{yu2026should}. 
WMs in robot learning, referred to as embodied world models, have gained traction because of the promise of an end-to-end system that can sense the environment, reason over action-conditioned dynamics, and perform a given task. 
Although vision-language-action models (VLAs), reinforcement learning (RL), imitation learning (IL), and model predictive controllers are also end-to-end techniques, the strength of a WM is, by definition~\citep{world-models, lecun2022path}, predicting multiple possible state evolutions without pre-specifying the state of the dynamic system.

\begin{figure}[!hb]
    \centering
    \vspace{-0.1cm}
    \includegraphics[trim=0cm 0cm 0cm 0cm, clip,width=\linewidth]{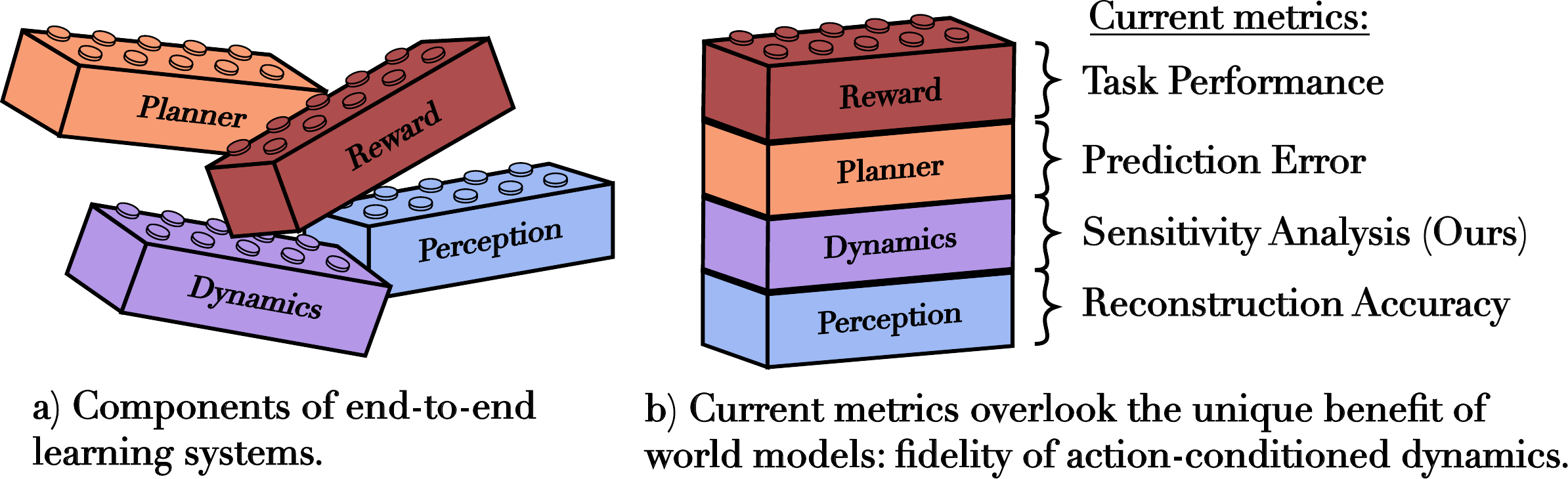}
    \vspace{-0.55cm}
    \caption{\textbf{The dynamics predictor of world models uniquely enables roboticists to audit end to end learning systems through observables alone.} 
    However, current metrics measure planning utility through one isolated component of the WM stack.
    Thus, new metrics that evaluate a system's dynamics fidelity of counterfactual actions are needed.
    }
    \label{fig:legos}
\end{figure}

Traditional learned dynamics models require specifying joint positions, object pose, friction, deformation modes, and other underlying mechanical properties. 
On the other hand, a full WM stack (Fig.~\ref{fig:legos}) learns an \textit{implicit} representation $p(z_{t+1:t+H} \mid o_{\leq t}, a_{t:t+H-1})$ directly from observations (e.g., pixels). Given the current state of the environment ($s_t$ if known, $s_t = o_t$ if from observations)

and a history of state-action-reward pairs $h_t = \{ (s_{t-H},u_{t-H},r_{t-H}),$ $... (s_t, u_t, r_t) \} $, a world model predicts the temporal response of the implicit state representation directly from raw observations. 

Counterfactual auditing~\citep{counterfactual-auditing}, i.e., the ``imagined" environment response under action sets $(u_t^1, u_{t+1}^1, \mathrm{...})$ vs. $(u_t^2, u_{t+1}^2, \mathrm{...})$, can be evaluated in terms of these observables. 
\uline{Inspecting counterfactual actions provides roboticists a window into \textit{auditing} the decision making of end-to-end learning systems which would otherwise be impossible without physical rollouts.} 

\begin{figure}[hbtp]
    \centering
    \vspace{-0.15cm}
    \includegraphics[trim=0cm 0cm 0cm 0cm, clip,width=\linewidth]{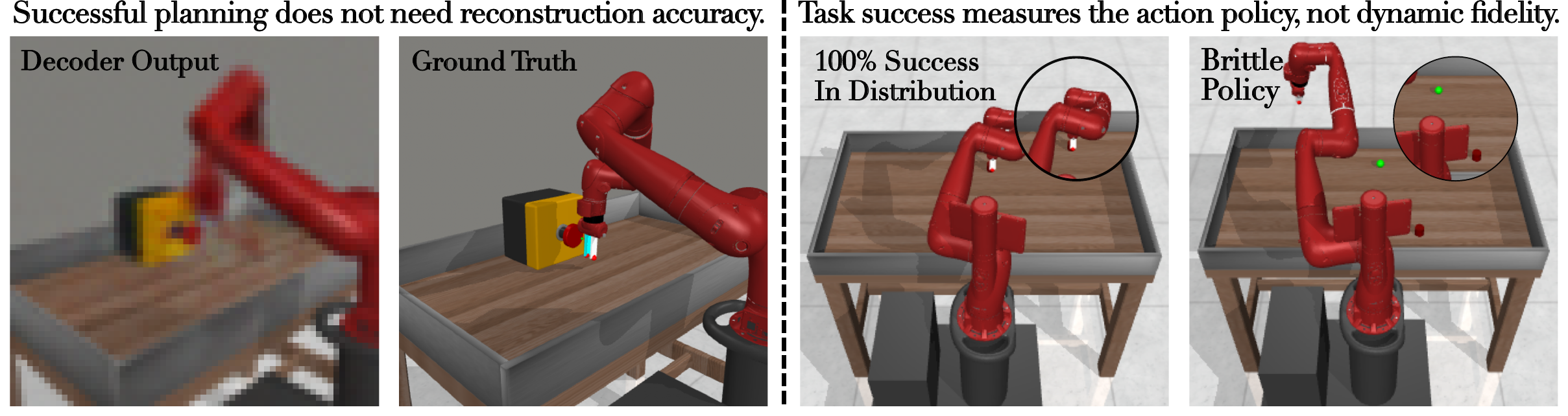}
    \vspace{-0.65cm}
    \caption{\textbf{Current metrics give false signals of planning utility.}
    Left: Visualizing ground truth interactions from an underfit decoder shows planning fidelity and reconstruction accuracy are distinct evaluations.  
    Right: An overfit, brittle action policy shows task performance and planning utility are distinct evaluations.
    }
    \vspace{-0.15cm}
    \label{fig:deceptive-metrics}
\end{figure}

Therefore, accurate dynamic predictions are a prerequisite for counterfactual auditing. 
Current WM metrics are ill-posed for verifying planning utility (Fig.~\ref{fig:deceptive-metrics}).
A world model with poor reconstruction capability can still be perfectly adequate for successful planning; in  model-based RL, low state reconstruction error does not necessarily translate to task performance \cite{mbrl-obj-mismatch}. 
Similarly, an overfit, highly brittle WM may have high task success within its training distribution but poor generalization. 

Auditability of embodied WMs requires both interpretability~\citep{interpreting-audit-xai}---model predictions are human interpretable---and actionability---the effects of alternative actions can be both imagined and executed.
Chain-of-thought semantic reasoning of modern VLAs~\citep{pi05, rt-2, open-vla, gr00t} is unreliable for robot trajectories~\citep{anthropic-cot, foutter2026vision};
although interpretable, language is a non-auditable proxy for predicting dynamic state evolution. 
WMs with latent transition models ~\citep{td-mpc2, daydreamer, vjepa2, leworldmodel} and World Action Models~\citep{open-wam} are best suited for counterfactual auditability because the learned dynamics models are both action-conditioned and interpretable via decoders.

\vspace{-0.3cm}
\section{Counterfactual Auditing Requires a Sensitivity Metric} \label{sec:method}
\vspace{-0.3cm}

Reconstruction fidelity, task success, and accuracy of dynamics predictions are all fundamentally independent objectives, each of which evaluates a different component of the WM (Fig.~\ref{fig:legos}b). 
Current evaluation metrics are directly correlated to training objectives, which jointly optimize perception, dynamics, and planning.
In order to measure the dynamics response under counterfactual actions, we need to measure the fidelity of specifically the predictive dynamics $z_{t+1:t+H}=f_\theta(z_t,u_t, h_t)$.

Evaluating latent dynamics models need not be different from evaluating standard dynamic models. 
Drawing from system identification and control theory, we treat world models as like any other input-output functions; validation requires testing the model's \textit{forced response} to known perturbations~\citep{yu2026should}, rather than trajectory or reconstruction error.

We use standard sensitivity analysis to model the environment's discrete impulse response $\Delta = f(s, u + \delta u) - f(s, u) \approx \frac{\partial f(s, u)}{\partial u} \delta u = J \delta u$. 
($f(s,u) = g\!\left(f_\theta(z,u)\right)$, where a learned linear probe $g$ maps the latent space $z$ to the state space $s$.) 
We measure the discrepancy between sensitivity of predicted dynamics $\hat{\Delta}$ and true dynamics $\Delta$ of a given trajectory as the Action Perturbation (AP) metric: $AP = {\lVert \hat{\Delta} - \Delta \rVert} / {\lVert \Delta \rVert} $. 
The AP metric is averaged over all evaluation trajectories, weighted by the true magnitude of impulse response.
Taking the difference of a difference cancels any constant state bias (i.e., poor modeling of $\partial f/\partial s$). 
After the one-step perturbation, the AP metric is reported over horizon $H=5$ with $\|\delta\|=0.2$; see Appendix~\ref{app} for more detail.
\enlargethispage{0.4\baselineskip}

While sensitivity analysis can also model response from perturbations in states ($\frac{\partial f}{\partial s}$) or even observations ($\frac{\partial f}{\partial x} = \frac{\partial f}{\partial q_\theta} \cdot \frac{\partial q_\theta}{\partial x}$ via chain-rule through the learned encoder $q_\theta$), at deployment, a robotic agent's capacity to interact with the environment is restricted to reasoning over alternative actions $u'$.
Although state sensitivity could measure stability or drift, the main advantage of WMs is the ability to evaluate counterfactual actions.
A successful controller requires high control sensitivity to generate diverse information gradients from rollouts scored on a task-centric cost.
In turn, a high accuracy on the prediction of control sensitivity determines whether a world model can be trusted in its decision making.
Thus, action sensitivity $\frac{\partial f}{\partial u}$ is the most important model response to audit.

\begin{figure}[hb]
    \centering
    \vspace{-0.25cm}
    \includegraphics[trim=0cm 0.2cm 0cm 1cm, clip,width=\linewidth]{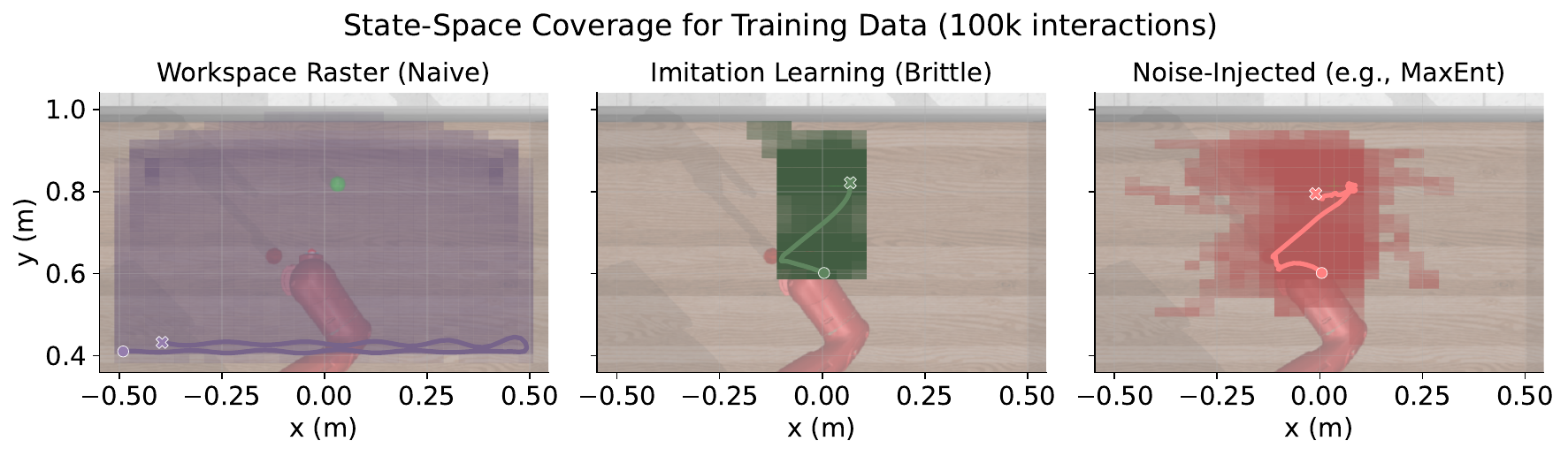}
    \vspace{-0.7cm}
    \caption{\textbf{SOTA models typically do not specify the coverage properties of the training data.} 
    Varied coverage strategies under a fixed budget (here, 100k interactions) drive model performance; metrics are evaluated on a fixed TD-MPC2 model~\citep{td-mpc2}, where only the dynamics predictor is retrained. 
    Left: Workspace raster dynamics head represents naive, uniform coverage.
    Middle: Imitation learning (IL) dynamics head represents a dataset with only successful demonstrations. 
    Right: Noise-injected dynamics head is trained on IL data with additive Gaussian noise. 
    }
    \vspace{-0.5cm}
    \label{fig:ee-coverage}
\end{figure} 

\section{Experiment \& Results}
\label{sec:result}
\vspace{-0.23cm}
\subsection{Evaluation Procedure}
\vspace{-0.23cm}
To isolate the dynamics fidelity of world models and measure how action sensitivity can vary under different data collection procedures, we finetune only the dynamics head with the standard dynamics consistency loss $\left\| \hat{z}_{t+1} - z_{t+1} \right\| $, where $\hat{z}_{t+1}$ comes from the dynamics predictor and $z_{t+1}$ comes from the encoded next state.
The dynamics prediction heads are frozen after training.
We freeze the released TD-MPC2 checkpoints~\citep{td-mpc2} for the other WM components; see Appendix~\ref{app:eval-dataset},~\ref{app:linear-train} for more detail.

Representative of the training distribution of state-of-the-art world models, we vary state coverage and finetune three representative dynamic heads on a fixed budget of 100k interactions (Fig.~\ref{fig:ee-coverage}):

\begin{enumerate}[itemsep=0pt, topsep=-4pt, parsep=0pt, leftmargin=3em] 
    \item Workspace raster: a standard yet naive data collection over a uniform grid of the workspace, which prioritizes state coverage over task-relevant actions. 
    \item Imitation learning: where training data comes from the expert rollouts of the released Metaworld RL action policy~\citep{mclean2025metaworld}, representative of popular diffusion-based methods~\citep{diffusion-policy}. 
    \item Noise-injected: similar to 2., but with varying scales of Gaussian noise injected into each action. From broad research consensus on the benefits of noise injection~\citep{laskey2017dart, sac-maxentrl}, we expect this model to be the best at sensitivity metrics, even at the slight expense of accuracy metrics. 
\end{enumerate}

With the perception encoder and evaluation data fixed, perception quality and task reward will be identical across the three dynamic heads.
We evaluate predicted state error and the AP metric over 50 \textit{fixed,} on-policy trajectories from released TD-MPC2 checkpoints for two sample tasks: 1) push a cube to a goal region (Push), and 2) insert a peg into the side of a fixed box (Peg).

\begin{figure}[hbtp]
    \centering
    \includegraphics[trim=0cm 0.1cm 0cm 0.1cm, clip,width=\linewidth]{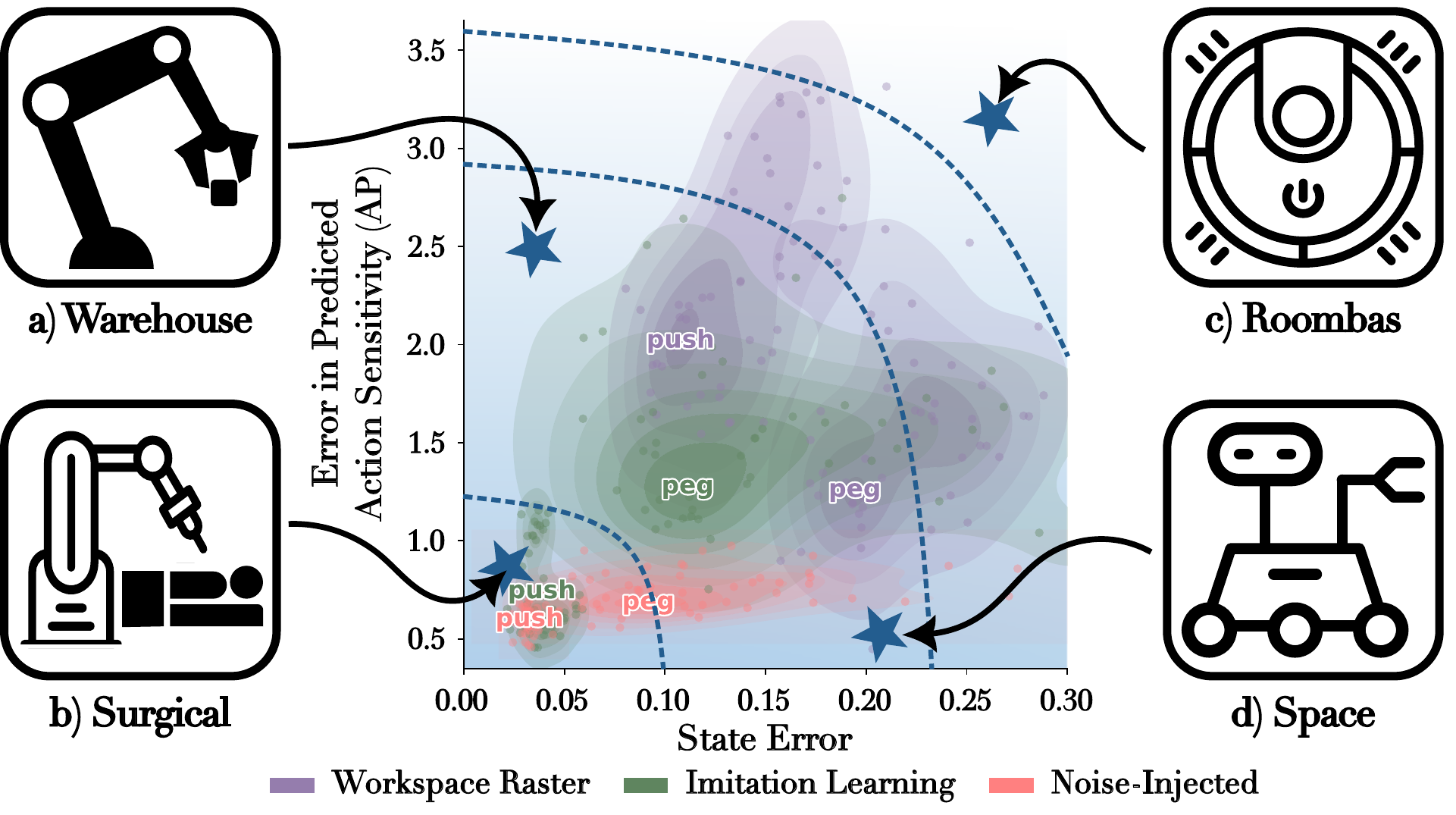}
    \vspace{-0.6cm}
    \caption{\textbf{Designing world models will often involve a tradeoff between low state error and low error in predicted action sensitivity.} For four example scenarios: 
    a) Structured environments, like open-loop pick and place, only require low state error. 
    b) High accuracy in both action sensitivity and state prediction is only possible when robot motion can be modeled and is repeatable. 
    c) Poor performance in both dimensions is acceptable with effectively unlimited time and compute for non-critical tasks.
    d) Action sensitivity is more relevant than state error in unstructured, unknown environments, where environmental uncertainty dominates error.
    }
    \vspace{-0.58cm}
    \label{fig:results}
\end{figure} 

\vspace{-0.22cm}
\subsection{Initial Results}
\vspace{-0.22cm}

Along the x axis of Fig.~\ref{fig:results}, the noise-injected dynamics and imitation learning head have similar state error accuracy; however, noise-injected clusters are lower on the y axis, with $\sim$16\% and $\sim$48\% better accuracy in predicting action sensitivity for Push and Peg respectively.
The noise-injected model is designed to have the highest probability of accurately modeling control perturbations, as additive noise to the action set during training is a heuristic to balance state exploration and policy exploitation.
The workspace raster dynamics model performs the worst across both metrics, likely because the limited, diffuse data is insufficient for accurate dynamics predictions.
Although the imitation learning model has low state error, the AP metric exposes the brittleness of the dynamics predictor when exposed to action perturbations from outside the support of its training demonstrations.

Ideally, all WMs are both accurate and have high control sensitivity, i.e., models in the lower left in Fig.~\ref{fig:results}.
However, until all environments are carefully controlled, robot motion is millimeter-scale precise, and learned models are exhaustive,
robot manufacturers and practitioners will need to make tradeoffs between low state error and highly accurate action sensitivity predictions. 
As illustrated by Fig.~\ref{fig:results} a)-d), one tradeoff may be favorable over another depending on if the environment is known vs unknown \textit{a priori}, robotic actions are precisely repeatable, and deployment conditions are structured vs unstructured. 
In carefully controlled environments where actions are repetitive, like open-loop pick-and-place tasks in warehouses, high state error is unacceptable, even at the cost of action sensitivity. 
Conversely, for robots in unknown, unstructured environments like space or underwater, state measurements are unreliable; instead, focusing on precise robotic hardware to accurately model the effects of actions is important.

\vspace{-0.35cm}
\section{Discussion}
\label{sec:discussion}
\vspace{-0.32cm}

World models provide decision transparency in end-to-end learning through counterfactual auditing, an otherwise impossible capability without physical deployment. Only WMs that \textit{accurately} capture the forced response of counterfactual actions can be reliable auditors of end-to-end learning systems. 

Current metrics use state-based analysis to evaluate a model that is inherently action-conditioned, and are thus ill-posed for verifying planning utility. 
Moreover, these metrics focus on the perception or reward component of the WM stack, but fail to evaluate its core component: the dynamics predictor.
We propose an action sensitivity metric to measure the fidelity of the dynamics predictor over the action space.
System identification can provide a rigorous framework for practitioners to analyze in which environments state or action sensitivity prediction accuracy is needed. 

Currently, the AP metric requires precise resetting of the environment to compare the effect of various action perturbations from the same initial state, which is inherently limited to simulation. Future work would require modeling sensitivity analysis without access to more than one ground truth model response in order to feasibly deploy to hardware systems. Furthermore, the connection between action sensitivity and planning utility remains an open area of research.

\acknowledgments{
This research was supported by National Science Foundation Grant CNS-2237576. We thank James Avtges and other members of the Interactive and Emergent Autonomy Lab at Northwestern University for their feedback during the manuscript review process. AI was only used for code generation and editing; all figures and text are human generated.}

\bibliography{references}  

\appendix
\section{Appendix}
\label{app}

\subsection{The Action Perturbation (AP) Metric}
\label{app:ap}

The AP metric evaluates the fidelity of world models in predicting the consequences of counterfactual actions.
A nominal trajectory with actions $u_{t:t+H}$ is compared with a trajectory where the first action is perturbed $u'_t = u_t + \delta$. Both trajectories start from the same initial state $s_t$. The difference in these trajectories between the simulator and the learned dynamics $f_\theta$ is the AP metric.

The true impulse response $\Delta = f(s_t, u'_t) - f(s, u)$ is compared with the predicted model response. (Note that a prediction horizon $H>1$ is a prediction response, but not an impulse response. See Tab.~\ref{tab:ap_ft_push}.) As discussed in Sec.~\ref{sec:method}, $f(s_t, u_t) = g(f_\theta(z_t, u_t))$, where a learned linear probe $g$ maps from the latent $z$ to the state $s$. For a given dynamics head, perturbation magnitude $\|\delta\|$, and prediction horizon $H$, the AP score is a weighted average over the evaluation trajectory set:
\[
AP = \sum_{i=1}^{N} w_i \, \frac{\|\hat\Delta_i - \Delta_i\|}{\|\Delta_i\|},
\qquad w_i = \frac{\|\Delta_i\|}{\sum_{j=1}^{N} \|\Delta_j\|}.
\]
Thus, a counterfactual evaluation trajectory is weighted by the size of the true response of the perturbation; in other words, accurately modeling larger impulse responses is more important.
(The weights are also indexed by $j$ to $N=34,400$ for normalization.) Extending AP to report by state dimension, i.e., AP for robot states or object states, may be useful when it is important to better accurately model some states over others.

A lower AP score is better: $AP=0$ when every predicted response is exact, $AP=1$ when the perturbed action is ignored altogether ($\hat{\Delta}=0$), and $AP > 1$ is worse than ignoring the perturbed action altogether.
To separate the linear probe error from dynamics error, we also report the ``probe floor": AP calculated with \(\hat\Delta\) replaced by \(\Delta^* = g(q_\theta(o'_{t+H})) - g(q_\theta(o_{t+H}))\), where $q_\theta$ is the learned encoder, $o'_{t+H}$ and $o_{t+H}$ are the ground truth perturbed and nominal observations, and $g$ as the trained linear probe. 
$\Delta^*$ does not depend on $f_\theta$, and instead compares the model response by passing the encoded observation through the linear probe. $\Delta^*$ represents the best possible AP score given the learned probe $g$.

See Tables~\ref{tab:ap_ft_push} and~\ref{tab:ap_ft_peg} for AP scores over $\|\delta\| \in \{0.1, 0.2, 0.5, 1.0\}$ and $H \in \{1, 2, 3, 4, 5, 10 \}$.

\subsection{Collecting the Evaluation Dataset}
\label{app:eval-dataset}
We evaluate over 50 fixed, on-policy trajectories of varying seeds, over 100 timesteps from the released TD-MPC2 agent per task. No dynamics head was trained on this data.
With a maximum horizon of $H=15$, $100-15+(\mathrm{initial~ state})=86$ initial states are perturbed over. After each perturbation trajectory, the simulator is reset. We perturb each timestep along 8 directions drawn uniformly on the unit sphere in the 4D action space, each scaled to perturbation magnitude \(\|\delta\|\). 
The perturbation is applied only at the first action; the remaining actions \(u_{t+1:t+H-1}\) are then rolled out in an open-loop fashion. 
Thus, $N = 86 \times 8 \times 50 = 34,400$.

\subsection{Training the Linear Mapping}
\label{app:linear-train}
To report state error, we train a linear mapping $g(z)$ from the 512-dimensional TD-MPC2 latent space to the 7D environment's state space (robot end-effector xyz, gripper, object xyz). It is fit in closed form by ridge regression on encoder latents of real observations (and not on predicted dynamics latents). The same mapping is frozen and used across all dynamics heads. 

The mapping $g(z) = Wz + b$ is trained on a mixture of expert, noise-injected expert, on-policy TD-MPC2, perturb-and-recover, and workspace raster. The advantage of a linear mapping, where \(g(\hat z') - g(\hat z) = W(\hat z' - \hat z)\), is used to additionally constrain \(W\). $g$ has a RMSE of 0.015\,m and 0.009\,m on the push and peg experiments respectively.

\subsection{AP Scores Over Varying Perturbation Magnitudes $\|\delta\|$ and Horizons}
\label{app:varied-perturbs}

State error and AP score for the Push and Peg tasks are evaluated over 50 on-policy trajectories across varied perturbation scales $\|\delta\|$ and horizons $H$, with results in Tab.~\ref{tab:ap_ft_push} and Tab.~\ref{tab:ap_ft_peg}.
Metrics are reported for the Imitation Learning, Noise Injected, and Workspace Raster dynamics heads, as discussed in Sec.~\ref{sec:result}, all of which are finetuned from the released TD-MPC2 checkpoint over 100k interactions from their respective data regimes. The probe floor is an engineered best-case baseline, in which the model predicts ground truth nominal and perturbed latent states; this represents the lowest possible AP score for a given environment based on the error from the learned linear probe. 

Lower scores are better for both metrics. $AP=1$ if the dynamics predictor ignores the given action; therefore, $AP > 1$ (shaded in red) denote perturbation predictions worse than predicting the nominal action. $AP>1$ is most often caused by the dynamics head predicting a response that is orders of magnitude larger than the true perturbed response (typically by $2-5$x).

As expected, the noise injected model, in general, outperforms the other dynamics heads in predictive action sensitivity. The results are more prominent in the peg environment, indicating that action sensitivity and diverse data collection are both more important  in more complex environments. For the more complicated task, modeling fidelity from state error is ambiguous and brittle, whereas the AP metric is consistent. Surprisingly, the noise injected model performs poorly at AP at $H=1$ in Push, likely overpredicting the effects of small perturbations.

\begin{table}[h]
\centering
\small
\setlength{\tabcolsep}{4pt}

\caption{State error and AP score for the \textbf{Push} task on 50 on-policy trajectories, for $\|\delta\| \in \{0.1, 0.2, 0.5, 1.0\}$ and $H \in \{1, 2, 3, 4, 5, 10 \}$. Unreliable world models, characterized by predictive action sensitivity worse than simply predicting the unperturbed action, are shaded in red. Best models, if outside confidence intervals, are highlighted in bold.}
\label{tab:ap_ft_push}
\begin{tabular}{@{}lcc ccc c@{}}
\toprule
Metric & $\|\delta\|$ & $H$ & Imitation Learning & Noise Injected & Workspace Raster & Probe floor \\
\midrule
\multirow{6}{*}{State err.} & -- & 1 & $\mathbf{0.018 \pm 0.001}$ & $0.023 \pm 0.001$ & $0.049 \pm 0.002$ & -- \\
 & -- & 2 & $\mathbf{0.023 \pm 0.001}$ & $0.025 \pm 0.001$ & $0.067 \pm 0.003$ & -- \\
 & -- & 3 & $0.028 \pm 0.001$ & $0.027 \pm 0.001$ & $0.091 \pm 0.005$ & -- \\
 & -- & 4 & $0.034 \pm 0.002$ & $\mathbf{0.030 \pm 0.001}$ & $0.113 \pm 0.007$ & -- \\
 & -- & 5 & $0.039 \pm 0.002$ & $\mathbf{0.033 \pm 0.002}$ & $0.135 \pm 0.008$ & -- \\
 & -- & 10 & $0.064 \pm 0.005$ & $\mathbf{0.051 \pm 0.003}$ & $0.224 \pm 0.010$ & -- \\
\midrule
\multirow{24}{*}{AP} & \multirow{6}{*}{0.1} & 1 & $\mathbf{0.75 \pm 0.02}$ & \cellcolor{red!12}$1.31 \pm 0.04$ & \cellcolor{red!12}$1.41 \pm 0.07$ & $0.60 \pm 0.02$ \\
 &  & 2 & $\mathbf{0.41 \pm 0.01}$ & $0.45 \pm 0.01$ & \cellcolor{red!12}$1.04 \pm 0.08$ & $0.44 \pm 0.02$ \\
 &  & 3 & $0.48 \pm 0.01$ & $\mathbf{0.44 \pm 0.01}$ & \cellcolor{red!12}$1.45 \pm 0.11$ & $0.43 \pm 0.02$ \\
 &  & 4 & $0.63 \pm 0.03$ & $\mathbf{0.54 \pm 0.01}$ & \cellcolor{red!12}$1.88 \pm 0.13$ & $0.45 \pm 0.02$ \\
 &  & 5 & $0.73 \pm 0.05$ & $\mathbf{0.63 \pm 0.02}$ & \cellcolor{red!12}$2.24 \pm 0.17$ & $0.46 \pm 0.03$ \\
 &  & 10 & $\mathbf{0.97 \pm 0.06}$ & \cellcolor{red!12}$1.32 \pm 0.11$ & \cellcolor{red!12}$3.49 \pm 0.34$ & $0.48 \pm 0.03$ \\
\cmidrule(l){2-7}
 & \multirow{6}{*}{0.2} & 1 & $\mathbf{0.76 \pm 0.02}$ & \cellcolor{red!12}$1.31 \pm 0.04$ & \cellcolor{red!12}$1.42 \pm 0.07$ & $0.58 \pm 0.02$ \\
 &  & 2 & $\mathbf{0.41 \pm 0.01}$ & $0.45 \pm 0.01$ & \cellcolor{red!12}$1.04 \pm 0.08$ & $0.39 \pm 0.01$ \\
 &  & 3 & $0.47 \pm 0.01$ & $\mathbf{0.43 \pm 0.01}$ & \cellcolor{red!12}$1.47 \pm 0.11$ & $0.36 \pm 0.01$ \\
 &  & 4 & $0.62 \pm 0.03$ & $\mathbf{0.52 \pm 0.01}$ & \cellcolor{red!12}$1.91 \pm 0.14$ & $0.37 \pm 0.02$ \\
 &  & 5 & $0.72 \pm 0.06$ & $\mathbf{0.62 \pm 0.02}$ & \cellcolor{red!12}$2.28 \pm 0.17$ & $0.37 \pm 0.02$ \\
 &  & 10 & $\mathbf{0.97 \pm 0.06}$ & \cellcolor{red!12}$1.31 \pm 0.09$ & \cellcolor{red!12}$3.65 \pm 0.36$ & $0.37 \pm 0.03$ \\
\cmidrule(l){2-7}
 & \multirow{6}{*}{0.5} & 1 & $\mathbf{0.77 \pm 0.02}$ & \cellcolor{red!12}$1.34 \pm 0.04$ & \cellcolor{red!12}$1.45 \pm 0.08$ & $0.57 \pm 0.02$ \\
 &  & 2 & $\mathbf{0.41 \pm 0.01}$ & $0.45 \pm 0.01$ & \cellcolor{red!12}$1.06 \pm 0.08$ & $0.35 \pm 0.01$ \\
 &  & 3 & $0.46 \pm 0.01$ & $\mathbf{0.42 \pm 0.01}$ & \cellcolor{red!12}$1.48 \pm 0.12$ & $0.31 \pm 0.01$ \\
 &  & 4 & $0.60 \pm 0.03$ & $\mathbf{0.51 \pm 0.01}$ & \cellcolor{red!12}$1.93 \pm 0.14$ & $0.31 \pm 0.01$ \\
 &  & 5 & $0.71 \pm 0.05$ & $\mathbf{0.60 \pm 0.02}$ & \cellcolor{red!12}$2.33 \pm 0.18$ & $0.31 \pm 0.02$ \\
 &  & 10 & $\mathbf{0.94 \pm 0.05}$ & \cellcolor{red!12}$1.26 \pm 0.07$ & \cellcolor{red!12}$3.71 \pm 0.37$ & $0.30 \pm 0.02$ \\
\cmidrule(l){2-7}
 & \multirow{6}{*}{1} & 1 & $\mathbf{0.79 \pm 0.02}$ & \cellcolor{red!12}$1.36 \pm 0.05$ & \cellcolor{red!12}$1.46 \pm 0.08$ & $0.57 \pm 0.02$ \\
 &  & 2 & $\mathbf{0.41 \pm 0.01}$ & $0.45 \pm 0.01$ & \cellcolor{red!12}$1.05 \pm 0.08$ & $0.35 \pm 0.01$ \\
 &  & 3 & $0.46 \pm 0.01$ & $\mathbf{0.41 \pm 0.01}$ & \cellcolor{red!12}$1.46 \pm 0.12$ & $0.30 \pm 0.01$ \\
 &  & 4 & $0.59 \pm 0.02$ & $\mathbf{0.49 \pm 0.01}$ & \cellcolor{red!12}$1.92 \pm 0.14$ & $0.30 \pm 0.01$ \\
 &  & 5 & $0.69 \pm 0.05$ & $\mathbf{0.58 \pm 0.02}$ & \cellcolor{red!12}$2.32 \pm 0.17$ & $0.30 \pm 0.01$ \\
 &  & 10 & $\mathbf{0.94 \pm 0.04}$ & \cellcolor{red!12}$1.01 \pm 0.04$ & \cellcolor{red!12}$3.27 \pm 0.28$ & $0.31 \pm 0.02$ \\
\bottomrule
\end{tabular}
\end{table}

\begin{table}[H]
\centering
\small
\setlength{\tabcolsep}{4pt}
\caption{State error and AP score for the \textbf{Peg} task on 50 on-policy trajectories, for $\|\delta\| \in \{0.1, 0.2, 0.5, 1.0\}$ and $H \in \{1, 2, 3, 4, 5, 10 \}$. Unreliable world models, characterized by predictive action sensitivity worse than simply predicting the unperturbed action, are shaded in red. Best models, if outside confidence intervals, are highlighted in bold.}
\label{tab:ap_ft_peg}
\begin{tabular}{@{}lcc ccc c@{}}
\toprule
Metric & $\|\delta\|$ & $H$ & Imitation Learning & Noise Injected & Workspace Raster & Probe floor \\
\midrule
\multirow{6}{*}{State err.} & -- & 1 & $0.025 \pm 0.003$ & $0.025 \pm 0.003$ & $0.075 \pm 0.006$ & -- \\
 & -- & 2 & $0.047 \pm 0.007$ & $0.044 \pm 0.007$ & $0.127 \pm 0.008$ & -- \\
 & -- & 3 & $0.076 \pm 0.011$ & $0.066 \pm 0.010$ & $0.171 \pm 0.009$ & -- \\
 & -- & 4 & $0.111 \pm 0.016$ & $0.092 \pm 0.013$ & $0.199 \pm 0.009$ & -- \\
 & -- & 5 & $0.154 \pm 0.020$ & $0.120 \pm 0.016$ & $0.218 \pm 0.009$ & -- \\
 & -- & 10 & $0.342 \pm 0.030$ & $\mathbf{0.254 \pm 0.027}$ & $0.278 \pm 0.014$ & -- \\
\midrule
\multirow{24}{*}{AP} & \multirow{6}{*}{0.1} & 1 & $0.89 \pm 0.04$ & $\mathbf{0.65 \pm 0.04}$ & $0.98 \pm 0.05$ & $0.31 \pm 0.02$ \\
 &  & 2 & $0.63 \pm 0.04$ & $\mathbf{0.48 \pm 0.02}$ & $0.92 \pm 0.07$ & $0.20 \pm 0.01$ \\
 &  & 3 & $0.82 \pm 0.06$ & $\mathbf{0.54 \pm 0.02}$ & \cellcolor{red!12}$1.15 \pm 0.12$ & $0.19 \pm 0.01$ \\
 &  & 4 & \cellcolor{red!12}$1.07 \pm 0.09$ & $\mathbf{0.64 \pm 0.02}$ & \cellcolor{red!12}$1.34 \pm 0.15$ & $0.19 \pm 0.01$ \\
 &  & 5 & \cellcolor{red!12}$1.36 \pm 0.12$ & $\mathbf{0.76 \pm 0.03}$ & \cellcolor{red!12}$1.50 \pm 0.17$ & $0.20 \pm 0.02$ \\
 &  & 10 & \cellcolor{red!12}$2.14 \pm 0.28$ & \cellcolor{red!12}$\mathbf{1.42 \pm 0.10}$ & \cellcolor{red!12}$1.91 \pm 0.21$ & $0.22 \pm 0.03$ \\
\cmidrule(l){2-7}
 & \multirow{6}{*}{0.2} & 1 & $0.89 \pm 0.04$ & $\mathbf{0.65 \pm 0.04}$ & $0.98 \pm 0.05$ & $0.30 \pm 0.02$ \\
 &  & 2 & $0.63 \pm 0.04$ & $\mathbf{0.47 \pm 0.02}$ & $0.92 \pm 0.08$ & $0.20 \pm 0.01$ \\
 &  & 3 & $0.82 \pm 0.06$ & $\mathbf{0.53 \pm 0.02}$ & \cellcolor{red!12}$1.16 \pm 0.12$ & $0.18 \pm 0.01$ \\
 &  & 4 & \cellcolor{red!12}$1.08 \pm 0.09$ & $\mathbf{0.63 \pm 0.02}$ & \cellcolor{red!12}$1.36 \pm 0.16$ & $0.18 \pm 0.01$ \\
 &  & 5 & \cellcolor{red!12}$1.39 \pm 0.13$ & $\mathbf{0.75 \pm 0.03}$ & \cellcolor{red!12}$1.53 \pm 0.18$ & $0.19 \pm 0.01$ \\
 &  & 10 & \cellcolor{red!12}$2.23 \pm 0.27$ & \cellcolor{red!12}$\mathbf{1.43 \pm 0.10}$ & \cellcolor{red!12}$1.98 \pm 0.20$ & $0.20 \pm 0.02$ \\
\cmidrule(l){2-7}
 & \multirow{6}{*}{0.5} & 1 & $0.90 \pm 0.04$ & $\mathbf{0.65 \pm 0.04}$ & $0.99 \pm 0.05$ & $0.30 \pm 0.02$ \\
 &  & 2 & $0.63 \pm 0.04$ & $\mathbf{0.47 \pm 0.02}$ & $0.93 \pm 0.08$ & $0.19 \pm 0.01$ \\
 &  & 3 & $0.82 \pm 0.06$ & $\mathbf{0.53 \pm 0.02}$ & \cellcolor{red!12}$1.17 \pm 0.13$ & $0.17 \pm 0.01$ \\
 &  & 4 & \cellcolor{red!12}$1.10 \pm 0.09$ & $\mathbf{0.62 \pm 0.02}$ & \cellcolor{red!12}$1.38 \pm 0.17$ & $0.18 \pm 0.01$ \\
 &  & 5 & \cellcolor{red!12}$1.41 \pm 0.13$ & $\mathbf{0.74 \pm 0.03}$ & \cellcolor{red!12}$1.55 \pm 0.19$ & $0.18 \pm 0.01$ \\
 &  & 10 & \cellcolor{red!12}$2.23 \pm 0.24$ & \cellcolor{red!12}$\mathbf{1.41 \pm 0.09}$ & \cellcolor{red!12}$1.97 \pm 0.19$ & $0.19 \pm 0.02$ \\
\cmidrule(l){2-7}
 & \multirow{6}{*}{1} & 1 & $0.89 \pm 0.04$ & $\mathbf{0.64 \pm 0.04}$ & $0.99 \pm 0.05$ & $0.30 \pm 0.02$ \\
 &  & 2 & $0.62 \pm 0.04$ & $\mathbf{0.47 \pm 0.02}$ & $0.92 \pm 0.08$ & $0.19 \pm 0.01$ \\
 &  & 3 & $0.82 \pm 0.06$ & $\mathbf{0.52 \pm 0.02}$ & \cellcolor{red!12}$1.17 \pm 0.13$ & $0.17 \pm 0.01$ \\
 &  & 4 & \cellcolor{red!12}$1.09 \pm 0.09$ & $\mathbf{0.61 \pm 0.02}$ & \cellcolor{red!12}$1.37 \pm 0.17$ & $0.17 \pm 0.01$ \\
 &  & 5 & \cellcolor{red!12}$1.40 \pm 0.13$ & $\mathbf{0.73 \pm 0.03}$ & \cellcolor{red!12}$1.53 \pm 0.19$ & $0.18 \pm 0.01$ \\
 &  & 10 & \cellcolor{red!12}$2.16 \pm 0.20$ & \cellcolor{red!12}$\mathbf{1.36 \pm 0.08}$ & \cellcolor{red!12}$1.92 \pm 0.17$ & $0.18 \pm 0.02$ \\
\bottomrule
\end{tabular}
\end{table}

\clearpage

\end{document}